# The AI Revolution: Opportunities and Challenges for the Finance Sector

*Authors* –
Carsten Maple, *University of Warwick, The Alan Turing Institute*
Lukasz Szpruch, *University of Edinburgh*, *The Alan Turing Institute*
Gregory Epiphaniou, *University of Warwick*
Kalina Staykova, *University of Warwick*
Simran Singh*, Financial Conduct Authority*
William Penwarden*, Financial Conduct Authority*
Yisi Wen, *University of Warwick*
Zijian Wang, *University of Warwick*
Jagdish Hariharan, *University of Warwick*
Pavle Avramovic, *Financial Conduct Authority*

***Disclaimer***: *This report was authored by researchers at The Alan Turing Institute (the Turing) with contributions from staff at the UK Financial Conduct Authority (FCA). This work was supported by EPSRC Grant EP/X03870X/1 & The Alan Turing Institute. The funding supported a pilot secondment scheme for FCA staff to collaborate with the Turing to identify new insights on how emerging technologies could impact financial services.*

***Note***: *The report does not represent views or policies of the FCA, nor does it constitute financial or other professional advice.*

# Table of Contents

**Abbreviations**:

AI – Artificial Intelligence
API – Application Programming Interface
DARPA – The Defense Advanced Research Projects Agency
ES – Expert System
EU – European Union
FCA – Financial Conduct Authority
FINRA – Financial Industry Regulatory Authority
GAN – Generative Adversarial Networks
GDPR – General Data Protection Regulation
IDE – Integrated Development Environment
LDA – Latent Dirichlet Allocation
LLM – Large Language Models
LSE – London Stock Exchange
MDP – Markov Decision Process
MiFID – Markets in Financial Instruments Directive
ML – Machine Learning
NER – Named Entity Recognition
NLP – Natural Language Processing
NMF – Non-negative Matrix Factorisation
NMT – Neural Machine Translation
NN – Neural Network
PCS – Principal Component Analysis
RegTech – Regulatory Technology
RL – Reinforcement Learning
RPA – Robotics Process Automation
SME – Small and Medium-sized Enterprises
SMT – Statistical Machine Translation
SVM – Support Vector Machines
VAE – Variational Autoencoders
XAI – Explainable AI

# Executive Summary

This report examines Artificial Intelligence (AI) in the financial sector, outlining its potential to revolutionise the industry and identify its challenges. It underscores the criticality of a well-rounded understanding of AI, its capabilities, and its implications to effectively leverage its potential while mitigating associated risks.

In its various forms, from simple rule-based systems to advanced deep learning models, AI represents a paradigm shift in technology's role in finance. Machine Learning (ML), a subset of AI, introduces a new way of processing and interpreting data, learning from it, and improving over time. This self-learning capability of ML models differentiates them from traditional rule-based systems and forms the core of AI's transformative potential.

The potential of AI potential extends from augmenting existing operations to paving the way for novel applications in the finance sector. The application of AI in the financial sector is transforming the industry. Its use spans areas from customer service enhancements, fraud detection, and risk management to credit assessments and high-frequency trading. The efficiency, speed, and automation provided by AI are increasingly being leveraged to yield significant competitive advantage and to open new avenues for financial services.

However, along with these benefits, AI also presents several challenges. These include issues related to transparency, interpretability, fairness, accountability, and trustworthiness. The use of AI in the financial sector further raises critical questions about data privacy and security. Concerns about the 'black box' nature of some AI models, which operate without clear interpretability, are particularly pressing.

A pertinent issue identified in this report is the systemic risk that AI can introduce to the financial sector. Being prone to errors, AI can exacerbate existing systemic risks, potentially leading to financial crises. Furthermore, AI-based high-frequency trading systems can react to market trends rapidly, potentially leading to market crashes.

Regulation is crucial to harnessing the benefits of AI while mitigating its potential risks. Despite the global recognition of this need, there remains a lack of clear guidelines or legislation for AI use in finance. This report discusses key principles that could guide the formation of effective AI regulation in the financial sector, including the need for a risk-based approach, the inclusion of ethical considerations, and the importance of maintaining a balance between innovation and consumer protection.

The report provides recommendations for academia, the finance industry, and regulators.

For academia, the report underscores the need to develop models and frameworks for Responsible AI and the integration of AI with blockchain and Decentralised Finance (DeFi). It calls for further research into how AI outcomes should be communicated to foster trust and urges academia to lead the development of Explainable AI(XAI) and interpretable AI.

The finance industry players are advised to be cognizant of data privacy issues when deploying AI and to implement a robust 'human-in-the-loop' system for decision-making. Emphasis is placed on maintaining an effective governance framework and ensuring technical skill development among employees. Understand the systemic risks that AI can introduce is also emphasised.

The regulatory authorities are urged to shift from a reactive to a proactive stance on AI and its implications. They should focus on addressing the risks and ethical concerns associated with AI use and promote fair competition between AI-driven FinTech and traditional financial institutions. The report advocates for regulatory experimentation to better understand AI's opportunities and challenges. Lastly, fostering collaboration between regulators, AI developers, and ensuring international coordination of regulations are deemed pivotal.

These recommendations pave the way for the effective integration of AI in the financial sector, ensuring its benefits are optimally harnessed while mitigating the associated risks.

# 1 Introduction

This report aims to study the impact of artificial intelligence (AI) on the finance sector, focusing on its practical applications, challenges, and potential benefits for driving innovation and competition. As a high-level concept, AI is a broad field of computer science that focuses on creating models capable of performing tasks that typically require human-like intelligence, such as understanding natural language, recognising images, making decisions, and learning from data. These tasks encompass complex problem-solving abilities and human-like decision-making, which have been a subject of interest for researchers for over seven decades (Agrawal et al., 2019; Furman and Seamans, 2019; Brynjolfsson et al., 2021).

In recent years, there has been a significant increase in AI's practical applications across various industries, such as finance, healthcare, and manufacturing, thanks to advancements in computing power, data storage, and low-latency, high-bandwidth communication protocols (Biallas and O'Neill, 2020). One reason for AI's widespread adoption in sectors like financial services is its versatility (Milana and Ashta, 2021). A Bank of England and FCA survey in 2022 found that 72% of surveyed firms reported using or developing machine learning applications (Blake et al., 2022).

The increased use of AI in finance can be partially attributed to intense competition within the sector (Kruse et al., 2019). The term 'Fintech' has been coined to describe companies that use digital technologies in their services. Compared to traditional financial institutions, fintech companies leverage technology to offer innovative and user-friendly financial services, including mobile payments, online banking, peer-to-peer lending, and automated investment platforms. Since these are often more convenient, efficient, and affordable financial services for consumers, they may disrupt traditional financial services through heightened competition, innovation, and a focus on customer satisfaction. Although traditional financial institutions were initially reluctant to adapt to these changes, many are now investing in digital technology and collaborating with emerging fintech firms to stay competitive. For example, Deutsche Bank sought to invest in supply chain financing and establish a partnership to incorporate supply chain solutions and technologies into their offerings. They collaborated with Traxpay, a German fintech company that provides discounting and reverse factoring solutions to its corporate clients (Hamann, 2021). This partnership has allowed Deutsche Bank to become a prominent player in the global supply chain financing industry. Using AI, financial institutions can gain a competitive advantage by introducing innovative services and improving operational efficiency (Ryll et al., 2020).

The finance industry generates a vast and constantly growing amount of data, including daily transactions, market trends, and customer information. Such rich data can be harnessed to train AI algorithms and create predictive models, making it an ideal domain for AI applications (Boot et al., 2021). These applications can identify patterns and predict future trends in the market, enabling financial institutions to make more informed decisions about investments and other financial operations. Moreover, AI can also analyse customer behaviour and preferences, offering tailored recommendations and personalised services (Zheng et al., 2019). By leveraging AI applications, financial institutions can optimise operations, reduce costs, and provide better customer service. With the increasing volume of data generated by the financial industry, the integration of AI technology is expected to become even more prominent, leading to more sophisticated applications and further transforming the financial landscape.

Despite the potential benefits of AI in the finance sector, industry experts and academic research suggest that financial institutions have not been able to fully leveraged AI's potential (Cao, 2022; Fabri et al., 2022). This is partly because of the numerous challenges and pitfalls of developing and using AI models. One of the primary concerns for customers is the issue of data bias and representativeness, as improper use of AI can lead to discriminatory decisions (Ashta and Herrmann, 2021). Furthermore, over-reliance on similar AI models or third-party AI providers can worsen the situation, potentially leading to the exclusion of certain groups of customers from the entire market rather than just a single financial institution (Daníelsson et al., 2022). A comprehensive approach is needed to address these challenges, including transparent audits of AI models and data sources, regular audits to assess AI algorithms' accuracy and fairness, and engagement with customers to ensure that their concerns are being addressed.

As AI becomes increasingly popular in the financial sector, firms face challenges related to the explainability and interpretability of AI models (Fabri et al., 2022). Such challenges can lead to reputational damage and a reluctance to adopt AI applications. Firms' stakeholders, including customers, investors, and regulators, demand transparency and accountability in decision-making processes. Lack of interpretability can make it challenging for firms to identify and address errors or biases in their decision-making mechanisms using AI, resulting in legal and financial consequences. Hence, explainability and interpretability are critical factors for AI's responsible and ethical use in the financial sector (Fabri et al., 2022). Such challenges emerging due to the use of AI in finance highlight the importance of effective model governance and regulations to ensure ethical and responsible use of the technology (Ryll et al., 2020). Such measures not only increase consumer trust in AI but also help financial firms avoid negative consequences like legal liability, reputational damage, and the loss of customers. However, stringent regulations may impose a significant burden on firms seeking to implement AI systems, including additional costs related to data privacy, security measures, or hiring more staff to monitor and maintain the technology, these costs may deter some firms, especially smaller ones with limited resources, from adopting AI.

Regulators are also leveraging AI to enhance the efficiency and effectiveness of regulatory processes. Organisations such as the London Stock Exchange (LSE) and the Financial Industry Regulatory Authority (FINRA) are embracing AI as a means of improving their regulatory capabilities (Prove., 2021). For example, the LSE has partnered with IBM Watson and SparkCognition to develop AI-enhanced surveillance capabilities, while FINRA uses AI to identify and prevent stock market malpractices (Prove., 2021). Regulators must balance the benefits and risks of AI and ensure appropriate safeguards are in place to mitigate negative outcomes. The recent surge in publications has shed light on various opportunities, challenges, and implications of AI in the financial services industry (Bahrammirzaee, 2010; Cao, 2020; Hilpisch, 2020; Königstorfer and Thalmann, 2020).

This report seeks to complement and update previous surveys and literature reviews by achieving the following objectives: 1) summarising the key AI technology in finance services based on research from finance and information systems studies; 2) examining the benefits of AI use and adoption in the finance sector; 3) highlighting potential negative consequences and threats associated with AI use in the finance sector; 4) addressing and evaluating the challenges, 5) role of regulators in addressing these unintended outcomes while exploring the use of AI to enhance regulatory work; and 6) providing recommendations to academia, industry, and regulators.

# 2 Key AI Technology in Financial Services

AI models provide a set of tools and techniques that can enhance decision-making processes, reduce costs, and identify patterns that would otherwise require significant time and effort to uncover. This section aims to present an overview of the fundamental AI technologies used in the finance sector and delineate some of the principle applications, benefits and challenges of this technology for both researchers and practitioners in the field.

## 2.1 Machine Learning

Machine learning (ML) algorithms are developed to recognise patterns in data and make predictions or decisions without being explicitly programmed. In high-level classification, ML is classified into supervised, unsupervised, and reinforcement learning. Supervised learning models are trained on labelled data to predict an output variable based on input variables (Kotsiantis, 2007), but unsupervised learning involves discovering hidden patterns or structures in the data without any predefined output variables (Dougherty et al., 1995). Reinforcement learning is a relatively recent development in the field. It considers that an agent learns to make decisions based on the feedback from its environment in the form of rewards or penalties. This section will provide an overview of the three types of ML models.

### 2.1.1 Supervised machine learning

Supervised learning involves developing and training a model by providing a set of input-output pairs to learn a mapping function from the input to the output. After this phase, the model is tested and confirmed through a testing phase and can then be used to predict the most probable value of the target variable for unlabelled data. Several supervised learning algorithms are used, including well-known linear and logistic regressions. The other main techniques include the following. These techniques are used in a wide range of applications, such as financial prediction, portfolio management, credit evaluation, and fraud detection (Jemli et al., 2010; Krauss et al., 2017; Kvamme et al., 2018; Thennakoon et al., 2019; Zhong and Enke, 2019).

- *Decision trees*: A tree-based algorithm is used for both classification and regression tasks, which splits the data into smaller subsets based on the values of the input variables and decides based on these subsets.
- *Random forests*: An ensemble learning algorithm that combines multiple decision trees to improve performance and avoid overfitting.
- *Support vector machines (SVM)*: A classification algorithm that finds the best hyperplane to separate the data into classes.
- *Neural networks (NN)*: A multiple layers of interconnected nodes, each of which performs a simple computation on its inputs, inspired by the structure and function of the human brain, are used for both regression and classification tasks.

### 2.1.2 Unsupervised machine learning

The primary objective of unsupervised learning is to identify meaningful patterns or relationships within a set of input data. This can be achieved through various techniques, such as clustering, which groups similar data points together based on their similarity or distance from each other (Hofmann, 2001). Another major application of unsupervised learning is dimensionality reduction, where the algorithm finds a lower-dimensional representation of the data that still captures its essential features (Muscoloni et al., 2017; Saeys et al., 2016; Torshin and Rudakov, 2015). Additionally, unsupervised learning can be leveraged for anomaly detection, where the algorithm identifies data points that deviate significantly in the data.

Unsupervised learning is used in a wide range of applications, such as fraud detection, customer segmentation, portfolio optimisation, credit risk analysis, and market analysis (Bao et al., 2019;

Gomes et al., 2021; Kedia et al., 2018; Mittal and Tyagi, 2019; Umuhoza et al., 2020). Compared with supervised learning, unsupervised learning can be less interpretable as patterns and relationships discovered may not be straightforward and intuitive, and it is challenging to evaluate and validate the results of the analysis as there are no predefined labels or classes to compare against (Lee and Shin, 2020). Moreover, unsupervised learning can be more computationally intensive than supervised learning, given the more data and more complex algorithms required, and it can be more prone to overfitting as it does not have the same level of guidance or constraint as supervised learning. Some of the commonly used models of unsupervised learning include:

- *Principal component analysis (PCA)*: A technique used for dimensionality reduction that transforms high-dimensional data into a lower-dimensional space, while retaining as much of the original information as possible.
- *Association rule mining*: A technique used to discover relationships between variables in a large dataset, such as the method called Apriori.
- *Autoencoders*: A type of NN used for unsupervised learning, trained to reconstruct the input data. They are used for tasks such as image denoising and anomaly detection.
- *Generative models*: A class of models that can generate new data resembling the input data, such as generative adversarial networks (GANs) and variational autoencoders (VAEs).

### *2.1.3 Reinforcement learning*

The concept of reinforcement learning (RL) is that an agent interacts with an environment to learn how to maximise a reward signal. In this setting, the agent receives feedback from the environment in the form of rewards or punishments, and aims to learn a policy that maximises the cumulative reward over time, improving its performance through an iterative process of trial and error. The RL problem can be formalised as a Markov Decision Process (MDP), consisting of a set of states, actions, transition probabilities, and rewards (Arulkumaran et al., 2017). The agent learns a policy that maps states to actions and uses this policy to decide which actions to take in each state.

Reinforcement learning is less prevalent as compared with other ML approaches. However, its uniqueness led to its irreplaceability. It is particularly well-suited to problems with unknown optimal actions (i.e., correct answers). In financial services, it could be used for trading execution and dynamic pricing (Culkin and Das, 2017; Nevmyvaka et al., 2006; Yu et al., 2019; Zhang et al., 2020). RL algorithms can be grouped into model-free RL and model-based RL. Model-based RL entails indirect learning of optimal behaviour by constructing a model of the environment through observation of outcomes, including the next state and immediate reward resulting from actions taken (Abbeel et al., 2006; Levine et al., 2016; Levine and Abbeel, 2014). Model-based RL tends to learn with significantly fewer samples than model-free RL, while model-free RL is more generally applicable and easier to implement (Clavera et al., 2018). For model-free RL, two main approaches are policy optimisation and Q-learning:

- *Policy optimisation*: Enables an agent to learn the policy function that maps a state to an action without using a value function (Bennett et al., 2021; Yu et al., 2019).
- *Q-Learning*: Learns the optimal action-value function for a given MDP (Hasselt et al., 2016; Kumar et al., 2020). It works by iteratively updating the Q-values for each state-action pair, using the Bellman equation to estimate the expected reward for each possible action.

## 2.2 Expert System

An expert system (ES) is a type of AI system that imitates the expert decision-making abilities of a specific domain or field. ES utilises information in a knowledge base, a set of rules or decision trees, and an inference engine to solve problems that are difficult enough and require human expertise for resolution (Harmon P, King D, 1985). ES consists of three main components (Metaxiotis and Psarras, 2003; Sotnik et al., 2022):

- *Knowledge base*: It contains domain-specific knowledge and rules that the expert system utilises to solve specific problems. The knowledge base is typically created by domain experts and is organised to enable efficient access. The most used technique is the if-then rule.
- *Inference engine*: This expert system component uses knowledge in the knowledge base to draw conclusions and make recommendations. It utilises a set of rules or decision trees to guide its reasoning.
- *User interface*: This enables users to interact with the system, ask questions, and receive recommendations or advice. It mainly consists of screen displays, a consultation dialog and an explanation component.

Several factors differentiate expert systems from other mathematical models (Jackson, 1986). For example, (a) they can handle and process qualitative information; (b) inflexible mathematical or analogue methodologies do not restrict them and is capable of managing factual or heuristic knowledge; (c) their knowledge base of ES can be continually expanded with accumulating experience as necessary; (d) they can deal with uncertain, unreliable, or even missing data; and (e) they are capable of reflecting decision patterns of users.

ESs are used in various applications, such as financial prediction, credit risk analysis, and portfolio management (Bisht and Kumar, 2022; Mahmoud et al., 2008; Nikolopoulos et al., 2008; Shiue et al., 2008; Yunusoglu and Selim, 2013). They can be particularly useful in situations where the knowledge required is complex and difficult to acquire or where there is a shortage of human experts in a particular field. However, developing an expert system can be time-consuming and expensive, and the system's accuracy depends on the quality of the knowledge base and the rules used by the inference engine.

## 2.3 Natural Language Processing

Natural Language Processing (NLP) is a subfield of AI that enables computers to understand, interpret, and generate human language, thus facilitating natural communication between humans and computers (Mei, 2022; Ruffolo, 2022; Xu, 2022). NLP has many practical applications in financial services, such as chatbots for improving customer experience. It can help both regulators and financial firms by extracting relevant information and managing voluminous documentation efficiently (Mah et al., 2022). Some commonly used algorithms in NLP include:

- *Text Pre-processing:* This involves cleaning and transforming raw text data into a format suitable for analysis and quickly reducing the data space required. Common pre-processing techniques include tokenisation (breaking down a text into smaller chunks) (Forst and Kaplan, 2006; Habert et al., 1998; Webster and Kit, 1992), lemmatisation and stemming (allowing tracing back to a single word's root) (Chrupala, 2014; Yang et al., 2016).
- *Sentiment Analysis:* Useful in determining the emotional tone or sentiment in text by analysing text data. Both supervised and unsupervised algorithms can be used for sentiment analysis (Adamopoulos et al., 2018; Fang et al., 2014).

- *Named Entity Recognition (NER):* Identifies and classifies named entities (such as people, organisations, and locations) in unstructured text into a set of predetermined groups (McCallum and Li, 2003; Sang and De Meulder, 2003; Sazali et al., 2016). NER can be performed using rule-based approaches or machine learning models.
- *Topic Modelling:* This refers to identifying topics or themes in a corpus of text data (Blei, 2003; Sridhar and Getoor, 2019). Popular techniques for topic modelling include Latent Dirichlet Allocation (LDA) and Non-negative Matrix Factorisation (NMF).
- *Machine Translation:* It involves translating text from one language to another. Although this is not new to many people, the necessary level of quality is still not satisfied. The main challenge with machine translation technologies is to keep the meaning of sentences intact along with grammar and tenses (Khurana et al., 2023). Machine translation can be done using rule-based approaches or statistical machine translation (SMT) models, or more recently, with neural machine translation (NMT) models.
- *Speech recognition:* This can help covert spoken words into written text, which has been applied in various settings such as interviews and conversations. Symbolic, statistical or hybrid algorithms can support speech recognition.

## 2.4    Robotics Process Automation

Robotics Process Automation (RPA) is increasingly used in financial services to automate repetitive, rule-based human tasks. It aims to reduce operational costs, increase efficiency, reduce human errors, and enhance the customer experience (Driscoll, 2018; Gotthardt et al., 2020). Some of the common applications of RPA in the financial service include:

- *Account opening and closing:* RPA can automate the process of opening and closing accounts, including data entry, verification, and document processing (Romao et al., 2019).
- *Claims processing:* RPA can automate the process of claims processing, including data entry, verification, and validation, thereby reducing the time and cost involved in claims processing (Oza et al., 2020).
- *Fraud detection and prevention:* RPA can identify and prevent fraud by monitoring transactions, identifying anomalies, and alerting the relevant stakeholders in real-time (Thekkethil et al., 2021).
- *Customer service:* RPA can automate customer service processes, including query handling, complaint management, and resolution, thereby improving customer satisfaction and experience (Kobayashi et al., 2019; Lamberton et al., 2017; Willcocks et al., 2015).
- *Compliance and reporting:* RPA can automate compliance and reporting processes, including data collection, validation, and reporting, thereby reducing the risk of errors and improving regulatory compliance (Anagnoste, 2018; Radke et al., 2020; Sibalija et al., 2019).

# 3 Benefits of AI use in the Finance Sector

Researchers and practitioners have outlined numerous benefits of using AI in finance. However, they have cautioned that the realisation of these benefits depends on the scale of the organisation, with large financial organisations benefiting more than smaller ones (Ashta and Herrmann, 2021). This reflects financial institutions' different capabilities and resources and the scope of their services.

## 3.1 Improving Decision-making Process

One of the biggest benefits of using AI in the finance sector is improving decision-making related to credit assessment, lending, and investment. By harnessing the vast amount of data generated by financial institutions, AI models are being used to automate and augment decision-making. This allows for a more accurate, faster, and informed decision-making process. For example, AI-based models are increasingly used for automated decision-making in lending (Königstorfer and Thalmann, 2022), as they can significantly improve the credit risk assessment of a loan applicant due to their reliance on diverse, often non-traditional, data sets.

When AI and big data are used together, they can detect weak signals, such as interactions or non-linearities across explanatory variables, which seem to increase prediction accuracy compared to traditional creditworthiness metrics. This results in estimates of increased economic growth at the macroeconomic level. AI disrupts the banking industry because it allows for the utilisation of more varied types of data that can produce more accurate credit risk projections. AI can manipulate "big data" gathered due to consumer behaviour, the digitisation of customer contacts, and information made available through sources like social networks.

The effect that AI-based models (using large amounts of data produced by social media activity) might have on the quality of credit scores has been the subject of theoretical analysis by researchers (Wei et al., 2016). Researchers have concluded that they might backfire by causing strategic changes in the behaviour patterns of potential borrowers on social media sites. For instance, they can limit their social media connections or prioritise relationships with people in socio-professional groups, such as civil servants, who are less susceptible to losing their jobs. This indicates that AI-based scoring models intended to enhance prediction results may eventually lead to behavioural adjustments that perform well according to the selected indicators.

Empirically, there is still substantial disagreement over how AI may assist with financial decision-making. Lenddo and Big Data Scoring, two fintech companies whose major business is treating massive amounts of data using AI algorithms, predictably support this practise. Because they believe that the low likelihood of payback and potentially high loan risk will not even cover the evaluation costs, traditional banks may frequently choose not to evaluate the creditworthiness of small borrowers (Bazarbash, 2019). In this way, the use of AI and its capacity to handle a larger variety of data allows organisations like FinTechs to venture into territory that has, up until now, been uncharted.

Additionally, AI-based robo-advisors have been employed to improve or fully automate investment decision-making. For example, AI-based models can recommend investment portfolio construction and re-construction (Ahmed et al., 2021) and efficiently drive ESG investment targets (Ashta and Herrmann, 2021). Further, NNs have been deployed to estimate and recommend optimal investment strategies (Chen and Ge, 2021). Given their risk appetite, robo-advisors have also been utilised to help lenders decide on the most optimal P2P loan investments (Ge et al., 2021). Most lenders on P2P lending platforms such as Lending Club are relatively young and unexperienced. To avoid making bad investment decisions these platforms are increasingly relying on robo-advisors.

## 3.2 Automating Key Business Processes in Customer Service and Insurance

Financial institutions have benefited from automating key business processes using RPA algorithms (Wittmann and Lutfiju 2021). Such RPA are mainly used as robo-advisors to support customer services in retail banking and, more recently, in wealth management (Kruse et al., 2019). Such robo-advisors can offer automated financial planning services like tax planning guidance, opening a bank account, recommending insurance policies, giving investment advice, and many other essential financial services.

Banks can make easy wins in key areas such as untapped client segments, lower acquisition costs, stronger usage of existing products and services, and improved access and scale by adopting an AI-first approach to customer interaction (Mckinsey, 2020). Customers are asking for financial services to be delivered with a wider range of goods and services anytime, anywhere (Zeinalizadeh et al., 2015). Financial organisations can no longer ignore the extraordinary advantages of integrating and utilising Robotic Process Automation (RPA) solutions in their environment. Data collection enhances the user experience and offers numerous benefits to customers by creating the impression that AI interactions are on par with those of humans. By providing personal data, consumers can obtain personalised services, information, and entertainment, frequently for little or no cost.

Access to personalised services also suggests that users will benefit from the choices made by digital assistants, which successfully match preferences with accessible possibilities without subjecting users to the cognitive and affective exhaustion that can come with decision-making. To maintain their competitive advantage and boost profitability, banks are placing increasingly more strategic importance on RPA. The main advantage of using RPA services in retail banking is that it allows banks to operate around the clock, deliver cutting-edge services, and improve client experiences while increasing efficiency and accuracy (Villar and Khan, 2021). The sharing economy has developed to give consumers more power. Real-time analytics and messaging require the end-to-end integration of internal resources to fully utilise those potentials. Banks must modernise their IT architecture and analytical skills to acquire, process, and accurately analyse client data.

Similar automation is also often used in insurance, for example, in the pre-validation of pre-approved claims. Dhieb et al. (2019) propose an automated deep learning-based architecture for vehicle damage detection and localisation. Cranfield and White (2016) explain how insurance claims outsourcing and loss adjusting firm managed to implement RPA (Robotic, Cognitive robotic, AI), leading to a team of just four people processing around 3,000 claims documents a day. Thus, robo-advisors have been helping insurers collect information about claims and process the gathered information quickly.

## 3.3 Algorithmic Trading Improvement

Another key area where AI has been applied is trading, such as equity trading and, more recently, trading in the foreign exchange market. Algorithmic trading has proven to be more efficient than human traders as, relying on diverse, real-time data sets, it can factor in market anomalies and account better for price differences (Cohen, 2022). AI-models can also send better trading signals to human traders, as they can identify unexpected market trends within a limited time. Unlike human traders, such models do not rely on sentiments, which have been proven to cloud the judgement of human traders. Thus, AI models used for equity trading can be more efficient than human traders.

Recognising the potential of AI to perform high-frequency trading, many researchers and practitioners have tried to incorporate advanced AI models, relying on, for example, neural networks and fuzzy logic to advance algorithmic trading. There are many examples of these AI-based models, such as an AI-model based on a reinforcement learning algorithm that can improve stock trading (Luo et al., 2019), an NLP-based model to factor in investors' sentiments to predict stock trading returns (Martinez et al., 2019), and fuzzy logic models for predicting trends in financial asset prices (Cohen, 2022).

Recently, researchers have also become interested in studying the impact of social media data on stock performance. The proliferation of advanced AI and ML techniques has facilitated this research. For example, Valencia et al. (2019) developed a ML to analyse how Twitter data can predict the price movements of several cryptocurrencies, such as Bitcoin, Ethereum, Ripple, and Litecoin. Similarly, Wolk (2019) has shown that advanced social media sentiment analysis can be used to predict short-term price fluctuations in cryptocurrencies.

## 3.4  Improving Financial Forecasting

Forecasting models are the most widely developed AI-based models in the finance sector. These models utilise diverse sets of traditional and non-traditional data compared to traditional techniques. For example, Óskarsdóttir et al. (2018) use smartphone-based data in combination with socio-demographic data to forecast consumer loan default. By utilising a diverse data set, AI-based forecasting models produce better, richer insights within a limited amount of time, thus vastly reducing the time and cost of producing a forecast.

Various AI-based models predict companies' bankruptcy, loan defaults of small and medium-sized enterprises (SMEs), and stock price fluctuations (Ahmed et al., 2022). For instance, Sigrist and Hirnschall (2019) developed a model to predict the default of SMEs in Switzerland. Li and Mei (2020) utilised deep learning neural network with two hidden layers to predict asset returns, while Ruan et al. (2020) employed ML models to forecast stock market returns based on investor sentiment. Petrelli and Cesarini (2021) combined different artificial intelligence techniques to predict high-frequency asset pricing. Furthermore, credit risk forecasting models have also been developed by Sigrist and Hirnschall (2019).

AI models have also been used for insurance claim prediction. A claim is a request that the insurer's business pay for an occurrence that is covered by a policy, such as a car accident, a house fire, or a trip to the ER. An insurance claim is a request for reimbursement for unlucky occurrences like a car accident, medical emergency, or house fire. An insurance client may request an explanation of why their claim was refused by using AI to predict insurance claims (Rawat et. al, 2021). A first-party insurance claim must be made during an accident or other occurrence. An automobile insurance claim, for instance, can demand payment for property damage, personal injury, or accident-related medical costs. Both feed-forward and recurrent neural networks make up these neural networks. Annual claims are predicted using these methods. After training the test data, one may examine the association between the test data and standardised data. They also built model accuracy and stopping criteria into constructing these models.

## 3.5  Improving Compliance & Fraud Detection

AI-based models can help financial institutions achieve compliance faster, in real-time, and with limited resources (Ashta and Herrmann, 2021; Deshpande, 2020; Fabri et al., 2022). Such models enable real-time monitoring of data, which is paramount for the timely detection of suspicious activities and considered one of the main benefits of using AI for fraud detection.

AI can also make financial regulatory reporting more efficient by uncovering previously undetectable patterns (Kerkez, 2020). Trained on large sets of historic data, models utilising

ML techniques can find hidden fraud patterns by considering non-traditional financial data (Milana and Ashta, 2021). Reinforced ML will most likely be used to model unusual financial behaviour (Canhoto, 2020; Milana and Ashta, 2021). For example, AI-based models can recognise fraud patterns by studying annual financial statements to identify the risk of financial irregularities within an organisation (Wyrobeck, 2020). ML techniques can also be used successfully for identifying money laundering activities (Ahmed et al., 2022).

## 3.6 Reducing Illegal Insider Trading

Insider trading in the stock market involves trading based on non-public information, which can be legal or illegal. Insider trading is legal, providing that it adheres to specific regulatory guidelines. On the other hand, illegal insider trading occurs when trading is conducted based on non-public information, such as private, leaked, or tipped information, before it is made public (Varma and Mukherjee, 2022; Islam, 2018). This could include information on new product launches, quarterly financial status, or acquisition and merger plans.

AI can be used to detect potential insider trading around price-sensitive events. This is achieved through clustering to identify discontinuities in an investor's trading activity and identify small groups of investors that act coherently around such events, indicating potential insider rings (Mazzarisi et al., 2022).

Such AI-based systems can help financial industries and regulators stay ahead of insider trading and other financial crimes, ensuring that the financial sector remains transparent, fair, and free from illegal activities. Ultimately, using AI to detect and prevent illegal trading can help maintain the integrity of the financial markets and protect investors from potential losses.

## 3.7 Reducing Operational Costs

Scholars have pointed out that the adoption of AI by financial institutions can lead to a reduction in operational costs. For example, AI can reduce loan default rates because, due to robust credit score assessment, financial institutions and FinTech lenders can improve customer targeting (Königstorfer & Thalmann, 2020). Due to automation, AI can contribute to reducing costs in terms of, for example, compliance (Kerkez, 2020) and detecting financial fraud (Ashta and Herrmann, 2021). Similarly, using AI chatbots can help reduce labour costs (Patil and Kulkarni, 2019). Further, they can also improve speed in providing customer service (shorter time) and availability (a robot can run 24/7 and has no sick leave) (Wittmann and Lutfiju 2021).

AI can also be used to reduce the cost of transactions. By minimising the involvement of humans, AI can be utilised to increase the speed and efficiency of the payment process. By automating workflows, offering decision support, and applying image recognition to documents, AI can enable the straight-through processing of payments (Barclays, 2019). Typically, businesses may use RPA to automate low-value jobs to scale up the advantages through the number of transactions processed. However, the advantages naturally end at a certain point. AI systems can learn, foresee, and anticipate based on available knowledge and past data, whereas RPA systems can validate, analyse, compile, calculate, and orchestrate repetitive and rule-based activities. In order to cut costs and boost revenue potential, firms can deliberately design intelligent automation systems by taking a holistic view of end-to-end processes (Deloitte, 2020).

## 3.8 Improving Financial Inclusion

Millions of previously uninsured and underserved poor people are moving away from cash-based transactions and into formal financial services, where they can access a range of services like payments, transfers, credit, insurance, stocks, and savings (WorldBank, 2020). The problem of information asymmetry between financial institutions and individuals can be solved by providing digital financial inclusion through AI access to various social networks and online shopping platforms, which generate a wealth of personal data (Yang and Youtang, 2020). People might be able to access credit, save money, make deposits, withdraw, transfer, and pay for goods and services using a mobile device with AI intelligence. This enables those with modest incomes to obtain services that are not available to them through the traditional banking system (Van and Antoine, 2019).

### 3.9   Strengthening Cybersecurity Resilience

Financial organisations have been traditionally exposed to many cybersecurity attacks, which have recently increased. AI can be used successfully to further strengthen financial organisations' cybersecurity resilience. For example, AI techniques can provide better protection from social engineering attacks such as phishing. The primary use of NLP in cyber security for the insurance sector will be to promote interactions between humans and machines (Ursachi, 2019). Insurance companies may use NLP to search enormous databases for email conversations to detect the possibility of a phishing attempt. NLP can spot harmful behaviour patterns by monitoring all emails that enter the organisation's network (Mansour, 2020). For instance, if an email containing hazardous links, files, or malware were to infiltrate the networks, NLP could scan and analyse the email, delete it from the system, and ensure it has no negative effects on any important procedures or data.

### 3.10   Takeaways from the Insurance Sector

The necessity for profitability, financial regulation, and company competition all contributed to adopting AI in financial services (Akyuz and Mavnacıo, 2021). It has been noted that before using AI, a few important obstacles need to be considered. Information asymmetry is a serious issue because algorithms that receive insufficient data can produce inaccurate predictions with direct and indirect implications in financial engineering and decision-making (Jan, 2021). Any new technology can potentially expand the threat surface, raising the organisation's risks. Data privacy has grown to be a major problem regarding the usage of AI in the industry due to the extensive use of data for forecasts. Additionally, there is concern that substantial investments in new technology may drive up insurance costs, making it unaffordable for economically disadvantaged groups. AI makes data administration easier, which is its key benefit in the insurance industry. This might also be advantageous for the finance sector, given how data is processed for decisions in both industries.

Machine learning can be used to organise unstructured and semi-structured datasets. Scholars and data analysts can use datasets from different insurance companies. In the insurance sector, machine learning may be used to more accurately anticipate risk, claims, and consumer behaviour (Vandrangi, 2022). Artificial intelligence has also been used to power conversation interfaces that intelligently present clients with different types of information. These interfaces use current data, machine learning, and natural language processing. Chatbots receive natural language data from previous customer encounters, which an intelligent system evaluates and instantly uses to learn how to reply to users in text. In the insurance sector, artificial intelligence may be applied in various ways, including responsive underwriting, premium leakage, cost control, arbitration, litigation, and fraud detection (Pirilä et al, 2022). Strong artificial intelligence techniques are being incorporated into insurance data to handle this issue in great depth.

# 4 Threats & Potential Pitfalls

While AI can offer significant benefits for the financial industry, researchers and practitioners point out that the use of AI comes with many threats/potential pitfalls. Therefore, organisations, users and regulators must remain cognizant and vigilant of the potential drawbacks associated with using AI to ensure that this technology is utilised fairly and efficiently.

## 4.1 Explainability and Transparency of AI-based Models

The decision-making process of AI models is often compared to a "black box" that lacks transparency (Buckley et al., 2021; Milana and Ashta, 2021; Ryll et al., 2020). As a result, users are unable to comprehend how the system operates, makes decisions, and the underlying reasons behind those decisions. This creates a challenge in identifying errors and biases in the system, which may result in inaccurate and unjust decisions. Furthermore, the lack of transparency can constrain the ability to enhance the AI system over time. Without an understanding of its operations, it becomes difficult to identify areas for improvement or optimise performance, limiting the potential of AI to its fullest extent. Thus, one of the major pitfalls is incorporating transparency into AI solutions to unlock its full potential, including increased efficiency, accuracy, and cost savings.

Researchers have called for increased explainability of AI models to address this lack of transparency. However, while academic research is striving to develop explainable and interpretable AI models, incorporating explainability in AI models can result in lower efficiency (Adadi and Berrada, 2018) and higher costs when applying AI models (Fabri et al., 2022). Furthermore, it remains unclear among researchers and industry experts what constitutes a satisfactory explanation concerning explainable AI. The varying explanations required by different stakeholders further complicate the development and comparison of different explanations (Fabri et al., 2022).

These issues can have negative consequences for the financial industry in multiple ways. First, if the decision-making process of an AI system is obscure, it becomes challenging to detect and rectify errors or biases in the system, potentially resulting in flawed or biased decisions with detrimental outcomes. Second, if people cannot comprehend how an AI system arrived at its conclusions, they may be less inclined to trust the system, leading to limited adoption and effectiveness. Third, using AI may heighten systematic risk as more financial firms implement the same tools and algorithms (Daníelsson et al., 2022). In such cases, opacity hinders the proper modelling and monitoring of such risks, raising the likelihood of market crashes.

## 4.2 Fairness of AI-based Models

AI systems can replicate and amplify biases present in the data used to train them. Failing to conduct a comprehensive investigation of the data utilised to train AI models could result in outliers and spurious patterns in the data leading to AI models producing inaccurate and biased decisions that perpetuate existing biases and discrimination in society. Moreover, historical data largely used for AI and ML training have inherent limitations in fully representing the future, particularly when crucial extreme events are absent from the available financial data. This increases the likelihood of AI model failures during a crisis.

## 4.3 Lack of accountability for AI Output

One of the main pitfalls in using AI systems within financial organisations is the lack of accountability for AI output. This becomes particularly problematic when AI is employed to make critical decisions with important implications, such as assigning falsely bad credit scores, which can deny access to a loan. In cases where such AI-based critical decisions are made based on inaccurate training or biased and unrepresentative data, it is challenging to determine who is accountable for these decisions (Ashta and Herrmann, 2021; Fabri et al., 2022).

Machine learning techniques and associated artificial intelligence technologies use historical training data to determine how to respond in various circumstances. They frequently update their databases and educational materials in response to fresh knowledge. Two significant issues arise when attempting to raise awareness of these technologies, which must be considered. First, decisions are made automatically without human involvement, and mistakes cannot be tracked. Second, the justification for a decision's formulation might not always be clear to auditors (DRCF, 2022).

AI is used in a variety of processes, including damage assessment, IT, human resources, and legislative reform. AI systems can quickly pick up on petitions, policies, and changes made due to those policies. They are also quick to decide. This strategy raises concerns about security, social, economic, and political dangers and decision-making accountability. This further erodes trust in AI-based systems and reinforces the need for AI transparency and explainability.

### 4.4  De-skilling of employees in the financial sector

The development of advanced AI techniques, coupled with the increased availability of data, has resulted in a growing number of companies and individuals becoming attracted to AI and utilising it in their operations. However, excessive reliance on AI can present various risks. For instance, it can diminish human skills (Milana and Ashta, 2021) and discourage people from developing the necessary skills to make decisions independently. Researchers, for example, have pointed out that human skills related to financial forecasting, planning and decision support will soon be in less demand as financial organisations adopt more AI systems (Kruse et al., 2021). At the same time, in other areas within the finance sector, employees will be undergoing upskilling to train how to work more efficiently and safely with AI.

### 4.5  Job Displacement

Implementing AI on a large scale in the financial sector, particularly in commercial banks, will likely result in job displacement for many workers. As automation of routine tasks replaces human tasks, financial institutes will require fewer employees, with fewer recruitment drives and the potential for early retirements or even layoffs. This could lead to discontent among bank employees, resulting in productivity losses that could offset some of the gains from technological advancement (Juneja, 2021).

AI also has the potential to automate many non-routine tasks that humans currently perform. This could result in significant changes to labour demand, job polarisation, and inequality. For example, there may be a shift towards stronger relative employment growth in high-paid or low-paid occupations, depending on which AI automates non-routine tasks. This shift could lead to significant changes in the workforce and potentially exacerbate existing inequalities, leading to economic instability (WhiteHouse, 2022).

While there is a growing demand for AI skills in the finance sector in the UK, particularly in financial trading, projections over the next 5, 10, and 20 years indicate significant estimated net employment reductions. This raises concerns about job displacement and the need for financial institutions and policymakers to develop strategies to mitigate any negative consequences (BEIS, 2021).

## 4.6 Data Privacy Challenges

Based on the survey conducted by (Kruse et al., 2019), the financial service industry is apprehensive about losing control of their data, which is a valuable asset for their business. If financial institutions lose control of their data, it can lead to significant financial losses, legal liabilities, and damage to their reputation. However, collecting and storing a large amount of data can pose challenges to data protection. Hence, it is imperative to ensure that data is collected and processed in compliance with relevant data protection regulations (Lee and Shin, 2020), including the European Union's General Data Protection Regulation (GDPR) and other industry-specific regulations, and to implement appropriate security measures to safeguard sensitive data. If data protection issues are not addressed, AI technologies can impede adoption in the financial industry by eroding customer trust and confidence in financial institutions.

Further, it is tenable to argue that the enormous potential of technological platforms, which use risk prediction models based on machine learning algorithms, to obtain and analyse data from a variety of sources, such as internet searches, social media accounts, shopping, and purchase information obtained from credit card companies, is a potential threat to user/client privacy in the context of financial services. For example, consumers seeking auto insurance may not be aware of the information gathered about them or the methods used to utilise it as a basis for risk assessment (Riikkinrn et al., 2018). This data may be collected without the content providers' awareness and occasionally without their consent.

There is a chance that the data is inaccurate even though the claimant is not allowed to change it. Such a privacy violation might have detrimental effects on customers (Davenport et al., 2020). Some people may even deactivate their social media accounts if concerned that their online behaviour may increase their insurance prices. The lack of a time limit on using a person's information gleaned from a social media account or another source when assessing risk is arguably the most worrisome problem, becoming more prevalent as credit assessment models rely increasingly on social media data for their scores.

## 4.7 Systemic Risk

Although there is currently limited evidence, researchers and practitioners have warned that using AI can increase systemic risk in the finance sector (Danielsson et al., 2021). For example, algorithmic trading, which relies heavily on advanced AI techniques, allows an algorithm to learn and adapt its trading strategies independently. In unstable markets, this may lead to increased volatility, which, as financial markets are increasingly inter-connected, can create spillover effects and increase systemic risk (Svetlova, 2022). Researchers have also warned that the use of similar AI-based models, trained on largely similar type of financial data sets, can significantly increase the herding behaviour among human traders due to the similarity of the AI output, which can further destabilise the finance system (Svetlova, 2022).

## 4.8 High Cost of Error

Implementing AI can be very costly for organisations within the financial industry, with the added risk of significant financial losses in the case of errors. This is especially true for commercial banking, where loans can amount to millions. While humans have traditionally evaluated such loans, the rise of AI means that systems will increasingly play a major role, with humans in an ancillary position. If these systems make an error, such as disbursing a loan to a non-creditworthy counterparty, the bank will bear the consequences (Juneja, 2021).

# 5 Challenges

While the above section on pitfalls outlines some of the important issues that may stem from the continuous and wide-spread use of AI in the finance sector, various challenges associated with using AI would remain. If not addressed appropriately, these challenges may slow the adoption of AI-based systems in the finance sector.

## 5.1 Availability and Quality of Training Data

AI models need to be trained on a large amount of data. The more data the AI model has access to, the more accurate and reliable its predictions and decisions will be. By training on a vast amount of data, the AI model can learn from diverse examples, enabling it to identify more subtle patterns and relationships that would not have been possible with smaller datasets. Additionally, using a large dataset helps to reduce the risk of overfitting, where the AI model becomes too specialised in the training data and fails to generalise to new data. Therefore, a large amount of data is crucial for achieving high accuracy and avoiding overfitting during AI training.

Although the financial industry has access to more data compared to other industries, most of this data cannot be used to train AI models effectively. This is mainly because many established financial service providers have not fully digitalised their business processes. This results in insufficient amounts of digitally available data, which presents a challenge for adopting AI in the financial sector (BoE and FCA, 2022); Cao, 2022; Kruse et al., 2019; Milana and Ashta, 2021).

Data quality is also paramount when training AI models. If the data used is incomplete, inaccurate, biased, or inconsistent, it can negatively affect the model's performance and lead to inaccurate or unfair predictions. Ensuring data quality is crucial for deep learning and when data is more unstructured and sourced from multiple sources (Greenspan et al., 2016; Lee, 2017).

Another issue that erodes data availability and quality, particularly user data, for training AI models in the finance sector is the various data privacy requirements, which are stricter for financial institutions than for other industry sectors (Kruse et al., 2019). However, advanced AI techniques, such as federated learning, can allow AI models to be trained on user data without compromising user privacy (Ashta and Herrmann, 2021). Thus, financial organisations need to ensure that the development and deployment of AI models also safeguard user privacy requirements satisfactorily.

Therefore, it is essential to ensure that AI models' data are high quality, accurate, and representative of the real-world situations it is intended to model. This can be achieved by implementing data quality controls (Lee and Shin, 2020), such as data cleaning, validation, and profiling, to ensure that the data is accurate, consistent, and bias-free. By ensuring high-quality data, AI models can make accurate predictions and help organisations to achieve their goals more effectively.

## 5.2 Use of Synthetic Data in AI-models

Given the challenges in accessing relevant, high-quality data, financial organisations and regulators are increasingly turning to generate synthetic data. One of the key challenges associated with using synthetic data in AI is ensuring the quality of this data. The synthetic data must represent diverse and unbiased real-world scenarios to ensure accurate and reliable results. Developers must carefully account for data bias during generation to avoid biased models. Synthetic data can also be limited in its representativeness of real financial data, and outliers must be considered during the generation process to avoid compromising privacy (FCA, 2022).

Another significant challenge is ensuring security and privacy when undertaking synthetic data generation. Synthetic data generation techniques require real data as input, which poses a risk to consumers' privacy rights. Developers must comply with data protection laws to protect consumers' privacy and avoid infringing on their rights. Adequate measures must be in place to secure the synthetic data and prevent any unauthorised access or misuse of the data. Financial institutions must address these challenges to effectively leverage synthetic data to develop accurate and reliable AI models while ensuring the privacy and security of their customers' data (FCA, 2022).

## 5.3 Selecting the Optimal ML Model

Like the traditional statistical methods, no single AI algorithm is effective for all problems. Using an unsuitable algorithm can result in poor performance, inaccurate predictions, or even the inability to solve the problem. Moreover, as highlighted in section 2, there are instances where traditional methods outperform AI models in addressing specific issues. Therefore, selecting the appropriate algorithm involves comprehending the problem's nature and data characteristics and understanding the strengths and limitations of various algorithms. Organisations might employ different algorithms for different purposes based on the desired accuracy, interpretability, and dataset nature (Lee and Shin, 2020).

The use of RPA is also challenging as most financial institutions struggle to understand where to use RPA in their business. As listed in (Lamberton et al., 2017), targeting RPA at a highly sophisticated process is one of the top 10 common issues in failed RPA projects, leading to significant automation costs that could be spent on automating multiple other processes. Partially this is because most organisations do not completely understand the capabilities of bots or how they operate (Cooper et al., 2019). Another concern that deters these organisations from using RPA is the protection of business processes and the flow of information between different jurisdictions (Cooper et al., 2019).

## 5.4 Legacy Infrastructure

The expansion of Information Technology (IT) architectures has been ongoing since the 1980s; however, it has not been consistently updated. As a result, many IT architectures today, including outdated hardware and software systems, have become a burden, making it difficult and complicated to incorporate modern AI techniques into these legacy systems (Kalyanakrishnan et al., 2018). Consequently, this discourages the use of AI.

One reason for this challenge is that legacy systems may not possess the necessary processing power or storage capacity to effectively train and operate AI models (Ryll et al., 2020), which rely heavily on these resources. This can lead to longer processing times and reduced accuracy. Further, traditionally, financial organisations have gathered data in silos, that is, in isolated IT systems. Hence, organisations must first transfer the isolated data to shared data lakes to facilitate significant training and modelling activities.Such IT initiatives are complex, expensive, and time-consuming, which can further slowdown the adoption of AI in financial organisations.

Additionally, legacy systems may not be able to integrate with modern AI tools and platforms, limiting an organisation's ability to utilise the latest AI technologies and resulting in missed opportunities for innovation and reduced social welfare.

## 5.5 Lack of Appropriate Skills

Using AI also challenges financial organisations as most employees do not possess the technological expertise required to effectively operate AI systems (Kruse et al., 2019). Many AI systems require specialised programming, data analytics, and machine learning knowledge. Without these skills, employees may encounter difficulties comprehending how to properly use and interpret the outcomes produced by AI systems.

Furthermore, the rapid pace of advancements in AI technilogies can make it hard for employees to stay up to date with the latest trends and optimal techniques. Hence, organisations might need to invest in ongoing education and training programs to guarantee that their employees have the necessary skills and knowledge to use AI systems competently.

Moreover, adopting AI can change job responsibilities and roles (Culbertson, 2018; LinkedIn., 2017). Certain tasks might become automated, making them redundant, while others may require fresh skills or an alternative approach to problem-solving. Thus, it is crucial for organisations to anticipate and plan for these changes and to provide their employees with the essential support and training to adapt to their new roles and responsibilities.

## 5.6 Requirement of Better Agility and Faster Adaptability

Agility and adaptability are crucial for managing the risks associated with using AI in the financial industry (Thowfeek et al., 2020). As previously discussed, these risks include data bias, security and privacy concerns, and the opacity of AI models. Companies must be agile and adaptable in addressing these risks, which can have significant consequences for the business and its customers. Also, the use of AI can result in increased competition in the financial industry, as companies with advanced AI capabilities can quickly gain an advantage over those lacking such technology. To remain competitive, companies must be agile and adaptable in adopting and incorporating AI into their operations. Furthermore, using AI can lead to changes in how businesses operate and make decisions, which can require adjustments to existing processes and structures. Therefore, agility and adaptability are crucial for effectively leveraging the benefits of AI in the financial industry.

## 5.7 AI Model Development Challenges

While AI techniques have advanced significantly in recent years, financial organisations still struggle to develop accurate, well-performing AI models. Apart from the issues outlined above, some challenges relate to the nature of the AI techniques.

For example, NLP is used in sentiment analysis for predicting stock prices or generating trading signals from processing rich sets of financial text data (Osterrieder, 2023) presenting unique challenges related to language interpretability. While humans can easily detect the same words having different meanings by evaluating the context within which they occur, NLP models struggle with this task (Khurana et al., 2023). Similarly, humans often use different words to express the same idea, making it challenging to process the language and design algorithms. Homonyms are particularly problematic for question-answering and speech-to-text applications because they are not in written form (Khurana et al., 2023).

# 6 Regulation of AI and Regulating through AI

Given the rapid proliferation of AI-based systems in the finance sector and the threats/pitfalls they may create for individuals, organisations and society, regulators across various jurisdictions have been investigating how and to what extent they should regulate the use of AI in the finance sector. To better understand this emerging technology and its benefits, as well as its associated threats, regulators have also become increasingly interested in harnessing its innovation potential for regulatory work. This has led to the emergence of algorithmic regulation, which can be defined as "regulatory governance systems that utilise algorithmic decision-making" (Yeung, 2017, p. 5). Thus, regulators are currently facing two important issues: regulation of AI and regulating through AI (Ulbricht and Yeung, 2021; Yeung, 2017). To address this, regulators use regulatory testing of AI, which we explore in detail below.

## 6.1 Regulation Of AI

A fundamental aspect of good financial regulation is enhancing public trust by ensuring markets function well. Consumers should feel safe investing in financial offerings that suit them without the risk of being defrauded or even misinformed when making the investment. For regulators to continue maintaining consumer trust in the market in the context of a changing technology innovation landscape, there is a responsibility to be cognizant of how markets are evolving and keeping sight of risks that could emerge for consumers when adopting emerging technologies such as AI along the way to establish the right safeguards.

Digital technologies and data have disrupted entire industries and in many cases, have brought about products and business models that do not fit well with existing regulatory frameworks – particularly in finance, transport, energy, and health industries. Regulation in these industries is paramount to safeguard safety and quality standards to ensure the ongoing provision of critical infrastructures (OECD, 2019). It is now ever more difficult to know what, when, and how to structure regulatory interventions in a rapidly evolving technological landscape with immense disruptive potential (Fenwick et al., 2017).

Four main regulatory considerations arise from the growing adoption of new technologies:

- *Consumer protection:* What are the implications for consumers, especially regarding how their data might be used in the provision of offerings leveraging new technologies, as well as the risks around investing in the offering.
- *Competition concerns:* What are the implications for the diverse players in the market, especially smaller firms looking to compete with well-established tech firms that start providing financial services.
- *Market integrity:* What are the implications for financial stability, especially if many consumers start investing in risky, unregulated offerings without being subject to protections.
- *Operational resilience:* What are the implications for the financial market infrastructure, especially in operational disruption or large-scale cyberattacks within a rapidly growing dependence on technology. The Covid-19 pandemic showed the importance of ensuring operational resilience to protect consumers and market integrity.

With these considerations in mind, deciding on the scope of any AI regulation is not a simple task, as evident by the cautionary approaches undertaken by regulators around the globe. One of the challenges is related to the established principle of technology neutrality, which presupposes that legal rules "should neither require nor assume a particular technology" (Reed, 2007, p. 264). Already, however, certain regulations have focused specifically on AI. For example, while claiming to be technology-neutral, the European Union (EU) AI Act focuses on one specific class of technology – AI. Further, the act has been criticised for evading the technology neutrality principle by providing a too broad definition of AI as part of its scope, which currently encompasses AI techniques that are not considered to pose significant risks to consumers (Grady, 2023).

Other exiting regulatory initiates have also proven that it is not easy to regulate AI use in the finance sector, given that AI relates to a wide spectrum of analytical techniques applied across various finance areas. The rapid development of AI techniques also makes determining the right scope and timing of legislation problematic as regulators want to avoid overregulating, which may stifle innovative AI use in the finance sector and thus deprive us of AI-related benefits (see above).

### 6.1.1 Risk-based approach to regulating AI

Abiding to the technology neutrality principle includes focusing on the outcomes of the technology use rather than the technology itself. However, not all outcomes of the use of AI in finance should be regulated, as this can easily result in overregulation. The risk-based approach to understanding how and when to regulate AI, which the EU has promoted, allows us to adopt a more granular understanding of AI use in the finance sector and to associate those outcomes with different levels of risk. At the heart of this approach is the desire of the regulators to achieve an optimal balance between promoting AI's innovative use and ensuring fair use of AI. The risk-based approach to regulating AI allows us to move away from focusing on finding the precise legal definition of AI, which given its broad scope of techniques, applications and rapid development, is elusive and difficult to define, but rather focus on any negative, unintended outcomes that such AI use may cause. Such an approach, based on the principle of proportionality (higher risk comes with higher requirements), can help regulators avoid overregulating AI.

On the EU level, the EU AI Act has put forward this risk-based approach to regulating AI. In its current version, the proposal of the EU AI Act views predominantly credit scoring services as high risks, which will be subjected to strict requirements in terms of development, deployment, monitoring, and reporting, all of which will require human oversight and a high degree of transparency and explainability. Ultimately, the EU AI Act, which cuts across different industry sectors, is not expected to substantially impact other finance sector areas. However, the extent of its influence remains uncertain and subject to interpretation.

### 6.1.2 Existing regulation on the use of AI in the finance sector

It is important to note that the finance sector is one of the most heavily regulated industries. As such, robust legal rules are already in place that can address some of the challenges that AI can pose. For example, the EU's Markets in Financial Instruments Directive (MiFID) already provides robust rules concerning algorithmic trading. Further, various anti-discrimination laws forbid using statistics that severely denigrate protected characteristics by posing a serious threat of bias. This type of legal protection is illegal based on the Equality Act of 2010, which forbids insurers from utilising algorithms that may result in discrimination based on appearance and physical attributes. This is an undisputed and obvious point. Indirect discrimination may occur even though the algorithms used in the risk individualisation process are not designed to analyse physical attributes (Mann and Matzner, 2019). However, the actual outcomes of the individualisation achieved by the algorithms would be particularly harmful to people who have a protected attribute. This type of discrimination, sometimes known as "unintentional proxy discrimination," is widely believed to be inevitable when algorithms are used to look for relationships between input data and goal variables, regardless of the nature of these relationships (Prince and Schwarcz, 2019) For example, the programme would not purposefully discriminate against people based on their gender. Some proxies, such as the colour or brand of the car, on the other hand, may unintentionally reproduce biases or unintended outcomes that a person would not deliberately incorporate into the system.

Regulators, however, have continued to evaluate whether the growing use of AI in the finance sector can negatively impact consumer protection, competition, financial stability, and market integrity (Bank of England and FCA, 2022).

### 6.1.3 The need for Human-In-The-Loop

Despite the growing efforts of regulators, several legal scholars have stated that regulatory efforts under the form of legal frameworks, principles, laws, and other measures, may not be sufficient to ensure the fair use of AI in the finance sector (Buckley et al., 2021; Zetzsche et al., 2020). This is largely because AI systems are often conceived as black boxes, which makes their regulatory supervision challenging (Buckley et al., 2021; Zetzsche et al., 2020). Instead, legal scholars advocate that financial organisations that utilise AI should adopt strict AI internal governance policies by promoting personal responsibility of senior managers who are responsible for an organisation's AI-based systems. By adopting a personal responsibility, managers can demand more transparency and explainability (at least for high-risk areas) of the AI systems their organisations develop, deploy and use. This may also require independent AI review committees, as Zetzsche et al., 2020 suggested.

This drive towards more AI explainability could, however, lead to a potential clash between AI explainability as desired by senior managers vis-à-vis as preferred by regulators. Scholars have already reported such disparity between regulators (preferring a wider scope) and financial institutions (preferring limited scope) with regards to AI explainability (Kuiper et al., 2020). In particular, while AI explainability may be desired, it is often expensive and can come at the expense of the AI model's performance and accuracy. Thus, defining the optimal, desirable point of the AI model's explainability has remained challenging.

## 6.2 Regulating Through AI

The continuing adoption of AI in financial services drives debate among regulators and industry on the most appropriate approach to regulatory oversight. A key question concerns whether existing frameworks will suffice or new measures will be required (Bank of England, 2022). In order to contribute to these discussions, regulators must develop their expertise in AI (through algorithmic regulation). Moreover, AI has significant potential to improve the work of regulators too, so acquiring hands-on experience is an essential activity in modern regulatory practice. This section, therefore, explores aspects of AI use from the regulators' perspective.

### 6.2.1 The regulatory context

Scholars describe the role of the regulator as intentional attempts to manage risk or alter behaviour to achieve some pre-specified goals (Black, 2014). The control mechanisms are three core components of setting standards, gathering and monitoring information, and making interventions or implementing sanctions to align with the desired goals (Ulbricht & Yeung, 2019). To be able to oversee the safety and soundness of their regulatory environments, regulators need to be able to make decisions that are both timely and informed. With a move away from the prescriptive practices of regulation and a growing emphasis within regulatory regimes on targeting outcomes, there is a greater need for more and better information. These same information resources can also improve a regulator's ability to discover the low-probability but high-impact events that cause the most harm to markets and consumers (Black, 2014).

*6.2.2 The opportunity*

An important focus for regulators in using AI is on better *informed* decisions, while stopping short of *automating* decision-making. Improving analytical techniques, exploiting the speed, scale, and volume of modern information processing are key enablers. The evolution in maturity from descriptive to predictive and prescriptive analytics are helping move the time window of oversight from what has happened to what might happen to what action could be taken (Lepenioti, 2020). This supports a shift from reactive to proactive supervision, extending capabilities from the core of gathering and monitoring information to providing ever greater direction for action. As a result, the feasibility of taking preemptive action before any potential harm arises becomes more realistic.. These analytical methods depend on good quality data from improved data collection and processing activities. Increasingly, to pick up on leading indicators involves capturing more varied data – including less structured forms like text and image – from websites, open-source locations, and API interfaces, then structuring and organising the data for analysis. Automation capabilities also play an important role in strengthening the efficiency and effectiveness of regulatory workflows for the increasingly essential end-to-end data pipelines and for the scalable execution of business operations. The combination of established workflow and business rule technologies with the newer generation of analytical capabilities has the potential to extend the reach of regulatory process automation.

*6.2.3 The adoption of AI by regulators*

The adoption of technology by regulators to supervise markets has been increasing phenomenally. Such use of digital technology, including hardware and software, has been termed Supervisory Technology, or SupTech. While Suptech is used in many regulatory fields, finance is seen as a leader, and the development and deployment of technology in the financial regulation space are becoming more widespread and more sophisticated. This change has been fuelled by developments in AI that have significantly impacted financial regulation bodies' collection, processing and analytics functions. A typical example of AI-driven predictive analytics is using supervised machine learning to predict the risk of misconduct among financial advisors, with the findings being passed to supervisors for follow-up (FSB, 2020). Another example describes an unsupervised model that has been used to help regulators assess whether firms have categorised the risk levels of their customers appropriately (CCAF, 2022). A further example concerns using NLP and machine learning to compare filings against historical patterns to flag those that may be more likely to be problematic. In all these instances, the supervisor is provided with insights through analytics and then follows up with further investigation. The essence of this role of AI is to help regulators work out where to look (Toronto Centre, 2022). There is interest, too, in prescriptive analytics, but at present little sign of active use. This position may be more behavioural than technical in origin as it aligns with the consensus view that there has to be a human-in-the-loop. Elsewhere, there are numerous examples of regulators exploiting AI techniques to capture and process more granular, diverse, and timely data than can be used to provide insights that had not previously been possible (CCAF, 2022). These insights inform various regulatory activities, including risk scoring, triaging, monitoring misconduct, and detecting fraud. Collectively, these techniques are showing signs of improving the efficiency and effectiveness of regulatory activities, and as organisations with limited resources, regulators have much to gain.

*6.2.4 The horizon*

The trends for the future use of AI for regulation are already apparent. The move towards predictive supervision, in particular, appears to be established among leading regulators and will bring benefits to consumers and markets through quicker prevention of harm and regulators through more efficient targeting of supervisory resources. In forecasting future ways of working, looking at other domains for transferrable lessons is informative. Scholars of legal practice suggest that judges' role may evolve from predicting outcomes and making judgments to specialising more on the judgement itself while drawing increasingly on AI-developed predictions. We could see similar patterns in regulation whereby greater use of predictions by supervisors leads to more specialised decision-making roles (Legg, 2019). In healthcare, machine learning techniques applied to medical imaging are helping improve the effectiveness of clinical decisions by offering the potential to validate or even improve clinicians' diagnoses (Jussopow, 2022). Using AI to provide a second opinion on supervisors' analyses could be its parallel in regulation.

The rapid improvement in generative AI drawing on foundation Large Language Models (LLM) also offers significant potential for use by regulators. Considering the case of authorising firms, documentation gathered from a firm such as business plans, correspondence and reports could be passed into a pre-trained generative model that has learned to identify those characteristics: high risk, medium risk, or low risk. Similar techniques would be relevant to enforcement activities whereby generative AI could be used to summarise evidence for supervisors' review or further to highlight potential areas of concern. Recent research on the effectiveness of LLMs has shown an improving ability to exceed the passing score in examinations used to assess competence across a range of professional disciplines following training on practice materials (Nori, 2023). Such AI capabilities could offer ways for regulators to help firms better understand their responsibilities as regulated entities, whether by providing simpler summaries of sometimes complicated legal text or natural language interfaces that allow firms to ask questions interactively of a corpus of compliance obligations.

*6.2.5 The challenges*

The adoption of AI by regulators poses numerous challenges, which mirror those relevant to the industry. Predictive supervision techniques will inevitably increase as regulators learn lessons from leading practitioners. This will bring a risk that the efficiency of these predictive methods leads to a gradual rebalancing in the workings of regulatory decision-making. This may be mitigated by requiring a human-in-the-loop and ensuring specific accountabilities. However, while human oversight will help reduce some of these challenges, there will also be a need to determine the trustworthiness of AI-generated contributions, especially as models become increasingly complex. Effective techniques for explainability will therefore be increasingly important. In addition, there are risks around bias and discrimination, whereby historical data that is not sufficiently representative has a negative influence on outcomes, then imperceptibly perpetuates problems as it becomes the future training data. Further work will be required to understand these concerns and to develop practical tools and solutions.

AI has great potential to improve the efficiency and effectiveness of regulatory activities. Some of the most significant benefits can be achieved by improving data collection, processing, and analysis. As the intelligence aspect of AI becomes more pronounced, further benefits will be found in better supporting the higher-value work associated with regulatory decision-making. However, regulators must always be mindful of their oversight role and ensure that they exhibit appropriate behaviours in their use of AI.

## 6.3 Regulatory Testing of AI

In order to better understand the benefits, threats and challenges of regulating AI and regulating through AI, a common approach explored is regulatory testing of AI. To achieve this, regulators worldwide have begun implementing a 'test and learn' experimentation approach to learn about new technologies such as AI. This focus towards regulatory testing and learning as opposed to the historical 'wait and see' approach is driven by several factors, including the disruptive potential of new technologies, the need for a more responsive regulatory design, and the growing interest in innovative products and services. The examples provided in this section will largely be grounded in the context of financial regulation. However, many of the themes and principles can also be extrapolated to regulation in other sectors.

Testing environments such as test beds, living labs, and sandboxes have provided an avenue for evidence-based regulatory testing to support innovation and regulatory governance. Each testing environment has its own distinctive features that can support regulatory decision-making and learning by bringing in various stakeholders (Kert et al, 2022).

Digital Sandboxes are environments that provide a controlled space for experimentation, development, analysis and evaluation. There have been a plethora of Digital Sandboxes that regulators around the world have developed. These sandboxes have had various use cases and seen strong industry engagement. The Monetary Authority of Singapore (MAS) recently launched Project Guardian, a collaboration with the financial industry to explore the economic potential of asset tokenisation (representing assets through a smart contract on a blockchain). The FCA held a 3-week DataSprint, which convened 120 participants from across regulated firms, start-ups, academia, professional services, data scientists and subject matter experts to collaborate on developing high-quality synthetic financial datasets to be used by the members of the digital sandbox pilot (FCA, 2020). Members of the digital sandbox gain access to a suite of features such as readily accessible synthetic data assets; an Integrated Development Environment (IDE) with access to solution development tools; an observation deck to view the in-flight testing of solutions; a showcase page to examine solutions relating to different themes; an ecosystem of relevant stakeholders in order to facilitate solution development from both technological and conceptual angles; and an application programming interface (API) where vendors can list their solutions and APIs to encourage greater interoperability and foster a thriving ecosystem. To provide an example of the solutions that have emerged from the digital sandbox, the firm Financial Network Analytics developed a solution that uses NNs to establish the usual patterns of behaviour between organisations and individuals to highlight anomalies that can be used to detect fraudulent payments (FCA, 2021).

The digital sandbox is not to be confused with the FCA's regulatory sandbox. The digital sandbox is for firms at the early proof of concept stage, whereas the regulatory sandbox helps firms prepare to take their services to the market. The digital sandbox can be seen as a mechanism to support the early testing of emerging technologies using the development features and the datasets available and forms part of the experimentation wing of the regulator.

The digital sandbox is often used in conjunction with the FCA's TechSprints to provide participants access to datasets for solution development and validation. TechSprints are regulatory-led hackathons that facilitate collaboration among experts across and outside financial services to identify and develop solutions to key problem statements. These solutions often form proofs of concept that regulators and industry can explore and develop further. These initiatives form part of the FCA's test-and-learn approach to emerging technologies to understand its potential in addressing challenges and unveiling opportunities. It is geared more towards understanding the various possibilities in which emerging technologies can be harnessed to meet desirable outcomes instead of one that is immediately ready to implement and fit for purpose. Instead, it establishes the groundwork for new use cases to build upon by understanding the art of the possible. Some notable examples of TechSprints have been on 'Anti Money Laundering and Financial Crime', which shed light on the potential of Privacy Enhancing Technologies to facilitate sharing information about money laundering and financial crime concerns while remaining compliant with data security laws.

The two examples above, alongside other regulatory testing initiatives, have some common themes and practices that tend to underpin them.

Firstly, the approach is driven by the 'problem statement' or challenge as opposed to a preference for a particular technology or solution driving the process. As opposed to building a technological solution and finding ways to apply it, regulatory approaches tend to identify the problems first and start considering solutions that could potentially address those problems. Second, it is often an iterative approach with no 'correct answer'. Instead, there is a stronger focus on ensuring that consumer privacy and protection are kept at the core while exploring the possible ways solutions can address a particular problem statement. More importantly, it also acknowledges that emerging technologies may not offer the best solution out of a range of options in some cases. Third, it helps explore which technologies could lend themselves to immediate 'off the shelf' solutions and which ones would be a consideration for the future. This is especially relevant when understanding the implications of scaling up prototypes, developing an operational tool, and maintaining it over time. Fourth, there is a strong component of learning from other players' experiences beyond financial services, including other regulators (BEIS, 2020). Finally, there is a forward-looking aspect which is still very much grounded in existing tooling capabilities (BEIS, 2022), allowing regulators and other stakeholders to explore the trajectory for technologies concerning specific use cases and, consequently which areas could benefit from further policy guidance.

At its core, regulatory testing aims to understand how an uncertain future can impact specific outcomes within an industry. While new technologies bring more uncertainty about their implications, it is also worth noting that it is a challenge that has always arisen in response to any change, and it has been met with approaches that involve scenario analysis and hypothesis testing. As such, the principles underlying innovative approaches to testing new technologies remain fairly similar, even if the approaches taken might become more advanced as they iterate.

Increasingly, sandboxes for specific types of technologies are becoming more widely adopted. The EU AI Act is a recently proposed regulatory framework for AI that aims to promote the development and adoption of trustworthy and ethical AI systems while ensuring that these systems are developed and used responsibly and transparently. The AI Act includes several key provisions, including requirements for risk assessment, transparency, human oversight, and data protection.

A key element of the AI Act is the proposal for EU member states to set up national AI regulatory sandboxes to provide a platform for companies to test their AI systems in a controlled environment without facing the full burden of regulatory compliance. These sandboxes aim to encourage innovation while ensuring that AI systems are developed responsibly and safely (European Parliament, 2022).

Similarly, the European Commission has recently launched the European Blockchain Regulatory Sandbox for innovative use cases involving Distributed Ledger Technologies (DLT) in order to establish a pan-European framework to facilitate the dialogue between regulators and innovators for private and public sector use cases (European Commission, 2023).

These initiatives fall under the wider bucket of anticipatory regulation and involve engaging with stakeholders, monitoring trends and developments in the market, developing new regulatory frameworks and sandboxes to support emerging technologies, promoting collaboration between industry participants and regulators, and actively shaping the regulatory environment to promote innovation, competition, and consumer protection (OECD, 2020; Nesta, 2020).

In moving towards anticipatory regulation, regulators are increasingly becoming "market makers" rather than "market fixers" (Mazzucato, 2016). This concept was introduced by economist Mariana Mazzucato, who argues that regulators should take a more proactive role in shaping markets rather than simply responding to market failures or crises.

According to this concept, regulators should promote innovation and investment in key areas, such as green technologies, healthcare, and education, by providing the necessary infrastructure, funding, and regulatory frameworks to support these industries. This approach involves a greater emphasis on collaboration between regulators, industry stakeholders, and other actors in the market rather than relying solely on top-down regulation.

In understanding the strategic rationale for the regulator in expanding into the realm of tech exploration, the considerations need to be rooted in the regulatory objectives. Ultimately, the regulatory objectives drive and justify the undertaking of these initiatives. Most regulators have a mandate to protect consumers and enhance market integrity, with the UK FCA having a third objective to promote competition in financial services in the interest of consumers. In a rapidly changing landscape, with technologies like blockchain, AI, and quantum computing becoming increasingly disruptive while also posing many opportunities, there is a role for the regulator in keeping pace with these developments not just as an observer but as an active player in channelling the use of these technologies down the right and responsible avenues.

Anticipatory regulation is not just about becoming aware of risks and developments earlier but also carries a 'market-making' aspect with it. Beyond more formalised procedures of standards and legislation, the act of regulatory signalling in itself has a market-making component. Through initiatives such as a digital sandbox programme, or a TechSprint initiative, the regulator can signal that they would like to see more innovation in a specific area while actively providing guidance and policy steers. In signalling their appetite to encourage innovation and help provide the right environment for firms to optimise their development, regulators can actively shape the currents of innovation while learning about new technologies. As such, while regulatory innovation had started primarily serving a learning purpose, it can also become an influencing force.

In conclusion, the idea of regulators as "market makers" is particularly relevant in emerging technologies, such as AI, blockchain, and fintech. These technologies are rapidly transforming the financial industry and creating new opportunities for innovation, but they also raise important regulatory challenges, such as data privacy, cybersecurity, and consumer protection. As Ramos and Mazzucato (2022) note, while AI applications can improve the world, with no effective rules, they may create new inequalities or reproduce unfair social biases. Similarly, market concentration may be another concern, with AI development being dominated by a selected few powerful players. They recommend an "ethical by design" AI mission underpinned by sound regulation to shape the technology's potential in the common interest. In doing so, they note that the key is to equip the policymakers to manage how AI systems are deployed rather than always playing catch up.

As the technological landscape evolves, the role of the regulator and the parameters within which it operates will also become increasingly blurry. There will be a need for guidance from the regulator around best practices, standards, and ethical considerations concerning new technologies. Consequently, regulatory experimentation will only increase in the future and become more collaborative and data-led. Ultimately, if a regulation has to become more forward-looking, regulators will have to move from 'playing catch up' and more towards helping create the rules and parameters of the game.

# 7 Recommendations

## 7.1 Academia

Academia has a strong role to play in supporting the regulation of AI and the research and development of AI to support financial regulation. Key to this role will be the active engagement with regulators and industry, and there are many good examples to build upon. We recommend further initiatives to support collaboration, including cross regulatory-academic secondments to understand the ways of working and share learnings, as has been the case in the project that supported this report. Other recommendations for academia are:

- Develop models, frameworks and recommendations for Responsible AI, which address issues around fairness and accountability.
- Propose how we can integrate AI with blockchain and De-Fi, which can improve the efficiency of both technologies and help utilise their potential better.
- Behavioural and experimental finance researchers need to investigate how AI results and descriptions must be presented so that customers develop trust and finally perceive the product as attractive.
- Development of explainable AI and interpretable AI: the current status of Explainable AI (XAI) requires significant time to run and is expensive. The Defense Advanced Research Projects Agency (DARPA) has invested 50 Million USD and launched a 5-year research program on XAI (Turek, n.d.), aiming to produce "glass-box" models that are explainable to a "human-in-the-loop" without greatly sacrificing AI performance.
- Focus on developing AI-based models combining different AI techniques while factoring in human intelligence. Scholars have agreed that combining AI techniques can create more accurate models, strengthening trust in AI systems.
- Developing AI models which adequately address issues concerning explainability and transparency (see Milana and Ashta, 2021, for example)

## 7.2 Industry

Despite the outlined benefits of AI for the finance sector, industry reports, and academic studies argue that AI has been used only for implementing "small benefits" such as cost reduction and process optimization (PwC, 2020). More opportunities lie ahead, which can be pursued more productively if the challenges outlined above are overcome. In particular, we strongly recommend stronger engagement with academia and regulatory bodies, especially regarding emerging technologies, projects and applications and their uses. Knowledge sharing (with appropriate commercial and regulatory safeguards) will advance the market and society. We also make the following recommendations to financial organisations:

- Be aware of data privacy challenges when developing and deploying AI models. Be aware of the unintended consequences and potential pitfalls associated with using AI.
- Bring human-in-the-loop (intervention): this is vital for several reasons 1) Human discretion may reduce the machine's Type I and II errors (that is, False Positives and False Negatives, respectively). 2) Builds trust in machine learning models. 3) Ensures accountability for decisions. 4) Ensures adequate evidence exists to deliver consequential regulatory actions. 5) Process privileged information and decisions safely and securely.
- Ensure effective governance framework within organisations and on industry level (e.g., Model Risk Management and data quality validation): include effective assessment and reviews of ML models from development to deployment. Ensure technical skills training of employees, i.e., train employees how to use AI-based systems and be aware of AI ethics.
- Understand better the threats AI can bring regarding systemic risk to the financial systems.
- Ensure accountability, verifiability, and algorithms, data, and design process evaluation.

## 7.3 Regulators

Regulators should move from a reactive to a more proactive approach to understanding emerging technologies such as AI in terms of both opportunities and challenges. Such a proactive approach can help regulators understand how best to regulate them. Regulatory intervention can address the threats and challenges associated with using AI in the finance sector.

- Correct the most salient unintended consequences of the use of AI based on a risk-based approach (regulate strictly only high risks).
- Promote fair competition between FinTech using AI and traditional financial institutions.
- Strike a balance between AI overregulation and promoting AI development and use in finance.
- Understand better the opportunities and threats of AI through regulatory experimenting.
- Assess the opportunities and challenges of regulating through AI.
- Ensure customer protection: regulate both financial institutions and algorithm providers.
- Address ethical concerns surrounding the use of AI in the finance sector and consider customer perception and trust when developing regulations for AI use in finance.
- Foster collaboration between regulators and AI developers. This could build upon existing mechanisms, including the Bank of England's AI Public-Private Forum (AIPPF) or the Veritas initiative bringing together MAS and the financial industry in Singapore to strengthen internal governance of the application of AI.
- Develop a regulatory framework for data sharing that balances privacy concerns with the need for data sharing.
- Ensure international coordination and consistency of regulations for AI in finance.